\documentclass[runningheads]{llncs}
\usepackage[T1]{fontenc}
\usepackage{graphicx,verbatim}
\usepackage{booktabs}
\usepackage{xcolor}
\usepackage{colortbl}
\usepackage{multirow}
\usepackage{amsmath,amssymb,amsfonts}
\usepackage[colorlinks=true,
            linkcolor=blue,
            citecolor=blue,
            urlcolor=blue]{hyperref}
\usepackage{enumitem}
\AtBeginEnvironment{thebibliography}{%
  \setlist[enumerate]{label={[arabic*]}, leftmargin=*, itemsep=0pt}
}
\definecolor{bestblue}{HTML}{DEEBF7}
\definecolor{secondgray}{HTML}{F0F0F0}
\definecolor{bestgreen}{RGB}{198, 239, 206}   
\definecolor{secondyellow}{RGB}{255, 235, 156} 
\definecolor{headergray}{RGB}{217, 217, 217}   
\definecolor{bestblue}{HTML}{DEEBF7}      
\definecolor{secondgray}{HTML}{F0F0F0}    
\definecolor{aopblue}{HTML}{DDE9FF}       
\definecolor{asdpeach}{HTML}{FAE3D5}      
\definecolor{dicemint}{HTML}{E5F0E8}      
\definecolor{hd95lavender}{HTML}{E6E0F5}  
\definecolor{deltagreen}{HTML}{199935}
\begin{document}

\title{R2AoP: Reliable and Robust Angle of Progression Estimation from Intrapartum Ultrasound}
%

\author{%
  Yuanhan Wang\inst{1}$^\dagger$ \and  
  Yifei Chen\inst{1}$^\dagger$\and  
  Beining Wu\inst{2} \and  
  Mingxuan Liu\inst{1} \and  
  Xiaotian Hu\inst{1} \and  
  Chunbo Jiang\inst{2} \and  
  Yijin Li\inst{1} \and  
  Changmiao Wang\inst{3} \and  
  Feiwei Qin\inst{2}\thanks{Corresponding authors.\quad $\dagger$ Co-first authors.} \and  
  Qiyuan Tian\inst{1}\protect\footnotemark[1]
}

\authorrunning{Y. Wang et al.}
\titlerunning{R2AoP: Reliable and Robust Angle of Progression Estimation}

\institute{%
  Tsinghua University, Beijing, China 
  \and
  Hangzhou Dianzi University, Hangzhou, China \\
  \and
  Shenzhen Research Institute of Big Data, Shenzhen, China\\
  \email{qinfeiwei@hdu.edu.cn, qiyuantian@tsinghua.edu.cn}
}

 \maketitle              
\begin{abstract}
Accurate estimation of the Angle of Progression (AoP) from intrapartum transperineal ultrasound is critical for objective assessment of labor progression, yet remains highly sensitive to imaging noise, boundary ambiguities, and the geometric amplification of local segmentation errors. We propose R2AoP, a reliable and robust AoP estimation framework that integrates structurally informed segmentation and confidence-guided geometric modeling to achieve stable and reproducible measurements. A three-branch local-structure-enhanced backbone improves the delineation of the pubic symphysis (PS) and fetal head (FH), while confidence-weighted contour fitting explicitly suppresses the influence of unreliable boundary points in AoP computation. To further improve performance under heterogeneous acquisition conditions, we introduce a lightweight geometry-reliable test-time adaptation strategy as an auxiliary component, enabling stable inference without target annotations. Extensive evaluations on multi-center benchmarks demonstrate consistent reductions in AoP error and boundary metrics compared with state-of-the-art AoP methods. Our source code is available at \href{https://github.com/baiyou1234/R2AoP}{https://github.co\\m/baiyou1234/R2AoP}.


\end{abstract}
\section{Introduction}

Assessment of labor progression during childbirth is central to obstetric clinical decision making and directly affects delivery mode selection, operative vaginal assistance, and intervention planning \cite{BAI2026103960}. Transperineal intrapartum ultrasound enables real-time bedside acquisition of fetal head descent information \cite{https://doi.org/10.1002/uog.7521}, providing quantifiable evidence for labor progression assessment \cite{10.1007/978-3-032-11616-1_8}. Among these measures, the Angle of Progression (AoP), defined by the geometric relationship between the pubic symphysis (PS) and the fetal head (FH), correlates with delivery outcomes and is widely used in clinical practice and research \cite{10.1007/978-3-032-11616-1_10}. Since AoP is determined by the pubic symphysis axis and the tangent to the fetal head contour \cite{Benchmark2025}, automated computation requires robust PS/FH localization and accurate segmentation \cite{10.1145/3696409.3700214}. However, as a cascaded geometric measure \cite{ZhuShe_Bridging_MICCAI2025}, boundary errors can be amplified by geometric operations even with high segmentation accuracy \cite{10.1007/978-3-031-96318-6_4}, leading to angular errors and reduced AoP stability \cite{OU2024108501}. Therefore, automated AoP assessment requires not only strong segmentation performance but also consistent delineation of critical structural boundaries to ensure reliable geometric measurement under domain shifts \cite{wang2025smart}.

\begin{figure}[t]
    \centering
    \includegraphics[width=1.0\linewidth]{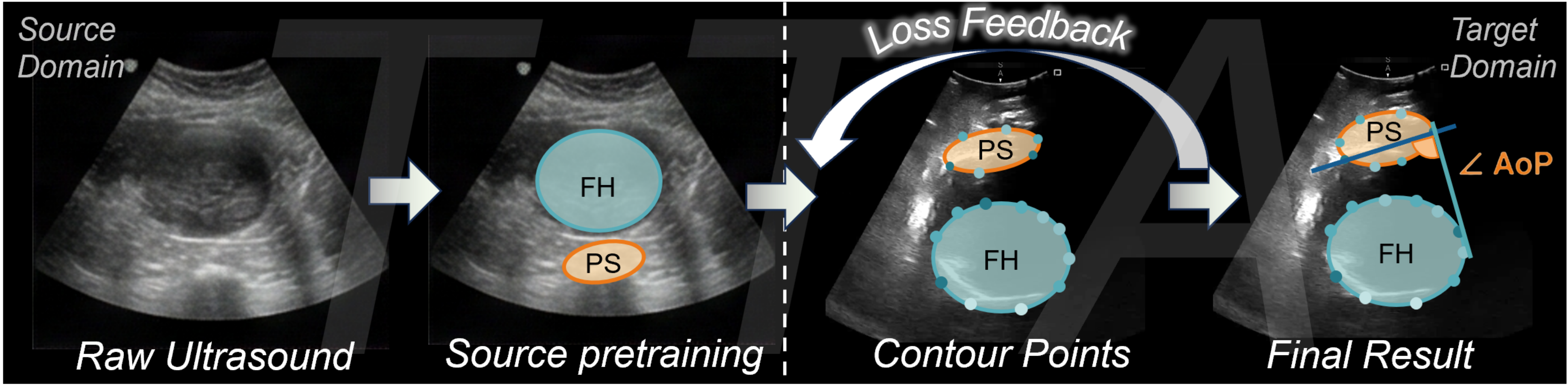}
    \caption{Illustration of the source-to-target pipeline of the proposed method. Source-domain training is followed by target-domain inference with geometry-reliable feedback for AoP computation.}
    \label{fig:Fig0}
\end{figure}

Intrapartum ultrasound imaging is commonly affected by speckle noise, low contrast, acoustic shadowing, and blurred boundaries \cite{chen2024dualpath}, leading to inter-subject and acquisition-related variability in PS and FH visibility and morphology \cite{CHEN2024107917}. The PS is slender, making it susceptible to noise and occlusion \cite{10.1007/978-3-032-11616-1_1}, while the FH contour may exhibit local discontinuities due to reflections or shadowing \cite{10.1007/978-3-031-96318-6_2}. These uncertainties expose PS/FH segmentation to topological disruption, fragmented false positives, and local adhesion \cite{10.1007/978-3-031-96318-6_3}, while boundary localization errors translate into systematic AoP deviations. Clinical deployment further involves domain shifts across devices \cite{Ravishankar_IEEETMI_2025}, acquisition parameters, and operators, where variations in models and scanning techniques alter image intensities and textures \cite{Ma_TTGA_2026}, degrading source-trained models in target domains \cite{ZHU2026104108}.

To address these challenges, this study proposes an integrated framework for automated AoP estimation that targets stable labor progression quantification under imaging degradation and domain shifts. The framework constrains AoP generation via structural awareness, cross-domain generalization, and output reliability modeling, reducing uncertainty and error amplification in geometric measurements. Unlike prior approaches that emphasize segmentation accuracy or assume consistent imaging, this work prioritizes cross-condition stability.

\section{Proposed Method}

\subsection{Baseline Network}
\begin{figure}[t]
    \centering
    \includegraphics[width=1.0\linewidth]{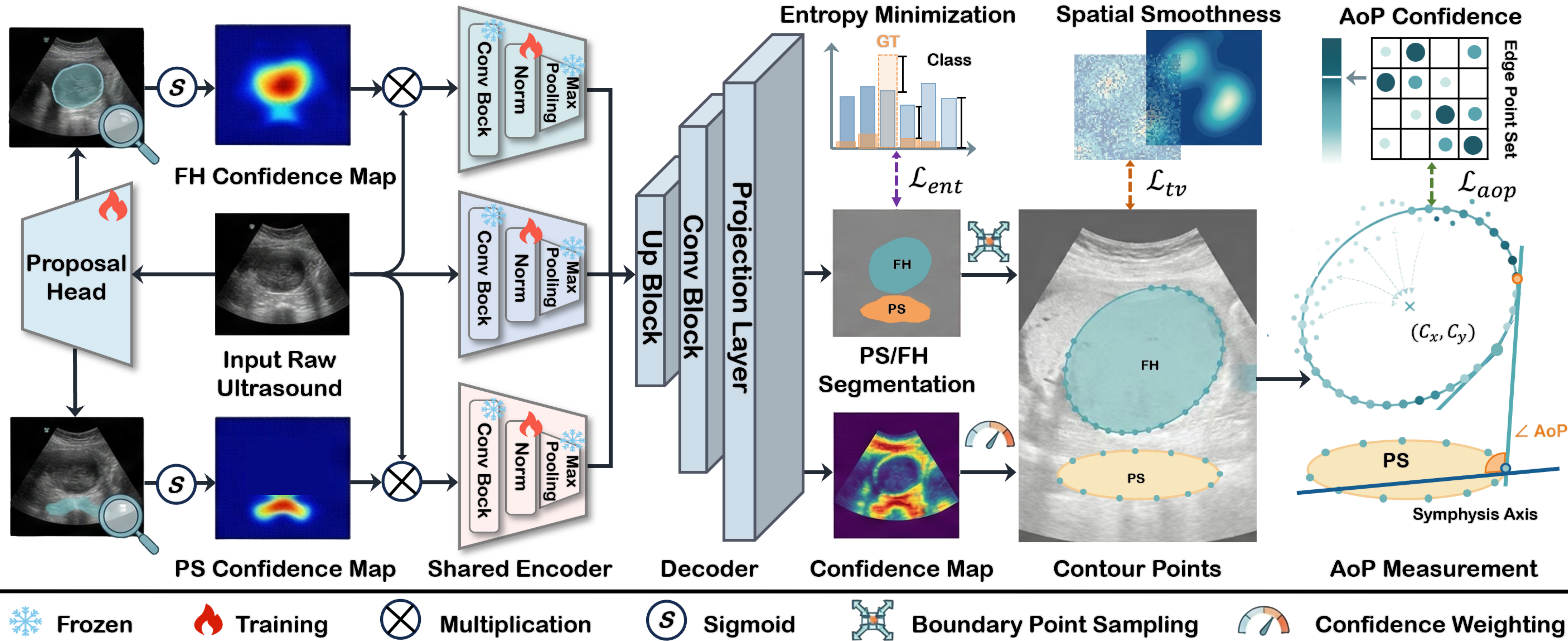}
    \caption{\textbf{The overall framework of the proposed method R2AoP.} The model jointly performs PS and FH segmentation, AoP measurement, and geometry-reliable test-time adaptation. Structural predictions are used for confidence-weighted contour fitting, geometric AoP modeling, and reliability-guided test-time adaptation.}
    \label{fig:placeholder}
\end{figure}

As illustrated in Fig.~\ref{fig:placeholder}, this study takes transperineal two-dimensional intrapartum ultrasound as input and jointly predicts PS and FH segmentation together with pixel-level confidence, based on which geometric estimation of AoP is performed. Key boundary points are extracted from the segmentation results and confidence maps to perform confidence weighted fitting of the fetal head contour, after which AoP is computed according to the geometric relationship between the pubic symphysis axis and the fetal head tangent, and the average confidence at critical boundaries is defined as the AoP confidence $C_{\mathrm{AoP}}$. During target domain testing, geometry aware test-time adaptation is introduced, where optimization objectives are constructed solely from unlabeled predictions and a small set of parameters is lightly updated under the geometric reliability constraint imposed by $C_{\mathrm{AoP}}$, thereby enhancing the stability and robustness of AoP estimation under cross-domain conditions.

\subsection{Three Branch Local Structure Enhanced Backbone} 
A two-dimensional U-Net is adopted as the segmentation backbone to produce pixel-level predictions of background, PS, and FH. To enhance the representation capability of critical local boundaries, a three branch shared encoding architecture is constructed, where one branch preserves global context and the other two emphasize complementary local region attention through a gating mechanism. Specifically, spatial gating maps are predicted on shallow high resolution features and multiplied with the original input in a pixel wise manner to form complementary views, after which the three streams share an encoder to extract multi-scale features that are fused at each scale and forwarded to the decoder to generate the final segmentation. The gating and fusion processes are learned end to end through the segmentation loss, thereby improving the accuracy and stability of PS and FH boundary prediction.
\subsection{Confidence Guided Geometric Modeling and AoP Estimation}
\label{sec:aop_conf}
Given the segmentation logits \(\mathbf{L}_{\mathrm{seg}}\), discrete labels are obtained as \(\hat{\mathbf{y}} = \arg\max \mathbf{L}_{\mathrm{seg}}\), after which the largest connected components of PS and FH are extracted and their boundary point sets are collected for downstream geometric analysis. To suppress noise and artifacts, the spatial confidence map \(\mathbf{S} \in (0,1)\) predicted by the network is used to weight boundary points, forming a confidence-weighted contour point set.

Based on the extracted contour points, ellipses are fitted to the PS and FH contours. The axial direction of the PS is defined as the major axis of its fitted ellipse, rather than the longest chord of the segmented region. From the major-axis endpoint adjacent to the FH, a tangent is drawn to the fitted FH ellipse, which together with the PS axis constitutes the geometric elements for AoP computation. Since the PS axis is derived from ellipse fitting, its endpoint may slightly extend beyond the segmented boundary when the predicted contour is irregular.

During geometric modeling, AoP is calculated in degrees according to the triangular relationship formed by the PS axis and the FH tangent:
\begin{equation}
\begin{aligned}
\mathrm{AoP}
&=
\arccos\!\left(
\frac{d_{13}^2 + d_{34}^2 - d_{14}^2}
{2 d_{13} d_{34}}
\right)
\cdot
\frac{180}{\pi},
\end{aligned}
\end{equation}
where \(d_{13}\), \(d_{34}\), and \(d_{14}\) denote the side lengths of the triangle determined by the PS axis and the FH tangent in the image plane.

To quantify measurement reliability, the AoP confidence is defined as the average spatial confidence of the key boundary sampling points involved in geometric modeling, where the boundary point set is denoted as \(\{p_j\}_{j=1}^{M}\), then:
\begin{equation}
\begin{aligned}
C_{\mathrm{AoP}}
&=
\frac{1}{M}
\sum_{j=1}^{M}
\mathbf{S}(p_j).
\end{aligned}
\end{equation}

Accordingly, the final outputs consist of the AoP value and its corresponding confidence \(C_{\mathrm{AoP}}\), which are subsequently used for geometry reliable constraint and quality assessment during test-time adaptation in inference.

\subsection{Geometry Reliable Constraint Based Test-Time Adaptation}
During target domain testing, no annotations are used. Unsupervised optimization objectives are constructed solely from predictions of current samples, and a small number of gradient updates is performed during inference to mitigate the uncertainty induced by domain shift without target supervision. Given the logits produced by the segmentation branch \(\mathbf{L}_{\mathrm{seg}} \in \mathbb{R}^{B \times 3 \times H \times W}\), pixel-level class probability maps are obtained via softmax:
\begin{equation}
p_{b,c,i,j}
=
\frac{\exp\!\left(\mathbf{L}_{\mathrm{seg}}^{b,c,i,j}\right)}
{\sum_{c' \in \mathcal{C}} \exp\!\left(\mathbf{L}_{\mathrm{seg}}^{b,c',i,j}\right)},
\qquad
\mathcal{C}=\{\mathrm{bg}, \mathrm{PS}, \mathrm{FH}\},
\end{equation}
where \(b \in \{1,\dots,B\}\) indexes samples in the batch, and \((i,j)\) denotes the pixel location with \(i \in \{1,\dots,H\}\) and \(j \in \{1,\dots,W\}\). For fixed \((b,i,j)\), the probabilities satisfy \(\sum_{c \in \mathcal{C}} p_{b,c,i,j} = 1\) after softmax normalization over classes.

\subsubsection{(1) Entropy Minimization Constraint}
To encourage the model to produce confident segmentation predictions in the target domain, a pixel-level prediction entropy minimization term is introduced as the core unsupervised objective:
\begin{equation}
\begin{aligned}
\mathcal{L}_{\mathrm{ent}}
&=
- \frac{1}{BHW}
\sum_{b=1}^{B}
\sum_{i=1}^{H}
\sum_{j=1}^{W}
\sum_{c=1}^{3}
p_{b,c,i,j} \log p_{b,c,i,j},
\end{aligned}
\end{equation}
where \(B\) denotes the batch size, \(H\) and \(W\) represent the spatial resolution, and \(\log(\cdot)\) indicates the natural logarithm. This term drives the predicted probability distributions from flat toward sharp, thereby alleviating boundary drift and ambiguous predictions under domain shift during test-time adaptation.

\subsubsection{(2) Spatial Smoothness Regularization}
To prevent entropy minimization from amplifying local noise, particularly under challenging clinical imaging conditions, a spatial smoothness regularization term is further imposed on the probability maps to enhance spatial continuity. A total variation formulation is adopted to penalize probability discrepancies between adjacent pixels:
\begin{equation}
\begin{aligned}
\mathcal{L}_{\mathrm{tv}}
&=
\frac{1}{BHW}
\sum_{b=1}^{B}
\sum_{c=1}^{3}
\sum_{i=1}^{H-1}
\sum_{j=1}^{W-1}
\Big(
| p_{b,c,i+1,j} - p_{b,c,i,j} |
+
| p_{b,c,i,j+1} - p_{b,c,i,j} |
\Big).
\end{aligned}
\end{equation}
This term encourages smooth and coherent probability distributions within anatomical structures, effectively reducing fragmented predictions.

\subsubsection{(3) Geometry Reliable Constraint}
In addition, the AoP confidence obtained in Section~\ref{sec:aop_conf} is incorporated as a geometry reliability constraint during test time optimization to improve robustness in practice under domain shift conditions. To promote geometric measurement stability, the following objective is defined:
\begin{equation}
\begin{aligned}
\mathcal{L}_{\mathrm{aop}}
&=
- \log \left( C_{\mathrm{AoP}} + \varepsilon \right),
\end{aligned}
\end{equation}
where \(\varepsilon > 0\) is a numerical stabilization constant. This term equivalently encourages increasing \(C_{\mathrm{AoP}}\), thereby constraining segmentation predictions through geometric usability under unlabeled conditions in target domain testing.

\subsubsection{(4) Overall Objective}
By integrating the above components, the overall test-time adaptation objective is formulated as a unified objective:
\begin{equation}
\begin{aligned}
\mathcal{L}_{\mathrm{tta}}
&=
\lambda_{\mathrm{ent}} \mathcal{L}_{\mathrm{ent}}
+
\lambda_{\mathrm{tv}} \mathcal{L}_{\mathrm{tv}}
+
\lambda_{\mathrm{aop}} \mathcal{L}_{\mathrm{aop}},
\end{aligned}
\end{equation}
where \(\lambda_{\mathrm{ent}}, \lambda_{\mathrm{tv}}, \lambda_{\mathrm{aop}} \ge 0\) denote weighting coefficients. To ensure lightweight and stable online updates, only the affine parameters of normalization layers and the parameters of linear layers are updated during testing, while all other network parameters remain frozen. For each test batch, a small number of gradient descent steps is performed to minimize \(\mathcal{L}_{\mathrm{tta}}\), enabling online domain adaptation in cross-domain scenarios under reliability-guided optimization.

\section{Experiments}
\subsection{Datasets and Preprocessing}
The PSFHS dataset \cite{chen2024psfhs} (source domain) is a publicly available intrapartum transperineal ultrasound pixel-level segmentation dataset, comprising 1,358 two-dimensional images from 1,124 pregnant women, with manual annotations of PS, FH, and background. PSFHS was split at the subject level into training and validation sets at a ratio of 9:1. All model selection and hyperparameter tuning, including the TTA settings, were performed solely using this internal split, while neither target-domain dataset was used during development. All images and labels were resampled to $256 \times 256$ using bilinear and nearest-neighbor interpolation, respectively.

The JNU-IFM dataset \cite{LU2022107904} (target domain 1) is an intrapartum transperineal ultrasound video dataset consisting of 78 sequences from 51 subjects. To reduce evaluation bias caused by correlations between adjacent frames, frames not simultaneously containing PS and FH were first excluded, and the temporally middle frame among the remaining valid frames was selected from each sequence, resulting in 64 external test images. The same preprocessing as the source domain was applied.

The IUGC 2024 challenge dataset \cite{bai2024intrapartum} (target domain 2) is evaluated using the official test set from the MICCAI Intrapartum Ultrasound Grand Challenge 2024, which contains 300 intrapartum transperineal ultrasound images with annotations for PS, FH, and background from diverse settings.

\subsection{Experimental details}
All experiments are conducted on a single NVIDIA A800 GPU, with the software environment consisting of Ubuntu 20.04.6 and Python 3.9. During source-domain training, the batch size is set to 8 and the number of training epochs is set to 30. The AdamW optimizer is employed with an initial learning rate of $1 \times 10^{-4}$ and a weight decay coefficient of 0.1. During test-time adaptation, all loss weights are set to 1.0. The adaptation learning rate is set to $1 \times 10^{-4}$, and one gradient update step is performed for each test batch. Only the affine parameters of normalization layers and the parameters of linear layers are updated.

\begin{table}[h]
\centering
\caption{Comparison with state-of-the-art methods on JNU-IFM 23 and IUGC 24 datasets. The best and second-best results are highlighted in \colorbox{bestblue}{blue} and \colorbox{secondgray}{gray}, respectively. $\Delta$ Gain denotes the performance improvement of our method relative to the second-best approach for each metric.}
\label{tab:comparison}
\resizebox{\textwidth}{!}{%
\begin{tabular}{c!{\vrule width 1.5pt}c!{\vrule width 1.0pt}ccc!{\vrule width 1.0pt}ccc!{\vrule width 1.0pt}ccc}
\toprule[1.5pt]
\multirow{2}{*}{\textbf{Method}} & 
\multicolumn{1}{c!{\vrule width 1.0pt}}{\multirow{2}{*}{\textbf{AoP$\pm$std}~$\downarrow$}} & 
\multicolumn{3}{c!{\vrule width 1.0pt}}{\cellcolor[HTML]{F0F0FF}\textbf{ASD$\pm${std}~(mm)}~$\downarrow$} & 
\multicolumn{3}{c!{\vrule width 1.0pt}}{\cellcolor[HTML]{E5F6FD}\textbf{Dice$\pm${std}}~$\uparrow$} & 
\multicolumn{3}{c}{\cellcolor[HTML]{F7F0F8}\textbf{HD100$\pm${std}~(mm)}~$\downarrow$} \\
\cmidrule[1.0pt]{3-5} \cmidrule[1.0pt]{6-8} \cmidrule[1.0pt]{9-11}
 &  & 
\textbf{PSFH} & \textbf{PS} & \textbf{FH} & 
\textbf{PSFH} & \textbf{PS} & \textbf{FH} & 
\textbf{PSFH} & \textbf{PS} & \textbf{FH} \\
\midrule[1.5pt]
\rowcolor{headergray}
\multicolumn{11}{l}{\textbf{Target Domain: JNU-IFM 23 Dataset \cite{LU2022107904}}} \\
\midrule[1.5pt]
DSTCT \cite{linguraru_intrapartum_2024} & $7.22_{\pm 6.27}$ & $17.87_{\pm 13.55}$ & $11.98_{\pm 7.35}$ & $20.53_{\pm 18.70}$ & $0.89_{\pm 0.09}$ & $0.82_{\pm 0.10}$ & $0.90_{\pm 0.10}$ & $70.22_{\pm 55.51}$ & $41.58_{\pm 23.39}$ & $63.74_{\pm 56.96}$ \\
IGT-Net \cite{CHEN2026128998} & \cellcolor{secondgray}$6.28_{\pm 6.88}$ & $16.96_{\pm 10.02}$ & $10.35_{\pm 7.17}$ & $19.67_{\pm 13.21}$ & $0.90_{\pm 0.05}$ & $0.84_{\pm 0.11}$ & $0.91_{\pm 0.05}$ & $77.39_{\pm 53.79}$ & $34.74_{\pm 20.81}$ & $75.09_{\pm 62.10}$ \\
UpNet \cite{BAI2025103353} & $44.92_{\pm 22.06}$ & $150.84_{\pm 32.94}$ & $125.68_{\pm 53.90}$ & $172.51_{\pm 36.89}$ & $0.23_{\pm 0.09}$ & $0.01_{\pm 0.01}$ & $0.23_{\pm 0.10}$ & $535.70_{\pm 112.59}$ & $212.69_{\pm 77.18}$ & $536.93_{\pm 109.95}$ \\
TransU-Net \cite{BAI2025103353} & $40.13_{\pm 26.16}$ & $64.22_{\pm 15.18}$ & $94.73_{\pm 48.00}$ & $94.74_{\pm 19.90}$ & $0.45_{\pm 0.12}$ & $0.02_{\pm 0.04}$ & $0.43_{\pm 0.12}$ & $317.90_{\pm 123.58}$ & $188.75_{\pm 60.07}$ & $341.38_{\pm 106.19}$ \\
AoP-SAM \cite{ZHOU2025125699} & $6.98_{\pm 5.28}$ & \cellcolor{secondgray}$5.35_{\pm 3.07}$ & \cellcolor{secondgray}$3.92_{\pm 2.60}$ & \cellcolor{secondgray}$5.92_{\pm 3.91}$ & \cellcolor{bestblue}$0.93_{\pm 0.04}$ & \cellcolor{bestblue}$0.87_{\pm 0.08}$ & \cellcolor{secondgray}$0.93_{\pm 0.04}$ & \cellcolor{secondgray}$21.15_{\pm 11.02}$ & \cellcolor{secondgray}$13.38_{\pm 9.23}$ & \cellcolor{secondgray}$18.65_{\pm 10.60}$ \\
USFM \cite{JIAO2024103202} & $11.95_{\pm 9.68}$ & $10.07_{\pm 7.95}$ & $7.53_{\pm 9.16}$ & $11.69_{\pm 10.67}$ & $0.87_{\pm 0.10}$ & $0.79_{\pm 0.12}$ & $0.88_{\pm 0.11}$ & $35.99_{\pm 29.72}$ & $28.02_{\pm 50.34}$ & $33.71_{\pm 30.10}$ \\
\textbf{R2AoP (Ours)} & \cellcolor{bestblue}$\textbf{6.21}_{\pm 5.18}$ & \cellcolor{bestblue}$\textbf{3.12}_{\pm 2.88}$ & \cellcolor{bestblue}$\textbf{2.00}_{\pm 1.20}$ & \cellcolor{bestblue}$\textbf{3.49}_{\pm 3.62}$ & \cellcolor{secondgray}$\textbf{0.92}_{\pm 0.04}$ & \cellcolor{secondgray}$\textbf{0.86}_{\pm 0.08}$ & \cellcolor{bestblue}$\textbf{0.93}_{\pm 0.04}$ & \cellcolor{bestblue}$\textbf{13.14}_{\pm 13.06}$ & \cellcolor{bestblue}$\textbf{6.76}_{\pm 4.04}$ & \cellcolor{bestblue}$\textbf{12.28}_{\pm 13.18}$ \\
$\Delta$ Gain & \textcolor[HTML]{199935}{$-0.07$} & \textcolor[HTML]{199935}{$-2.23$} & \textcolor[HTML]{199935}{$-1.92$} & \textcolor[HTML]{199935}{$-2.43$} & \textcolor[HTML]{199935}{$-0.01$} & \textcolor[HTML]{199935}{$-0.01$} & \textcolor[HTML]{199935}{$0.00$} & \textcolor[HTML]{199935}{$-8.01$} & \textcolor[HTML]{199935}{$-6.62$} & \textcolor[HTML]{199935}{$-6.37$} \\
\midrule[1.5pt]
\rowcolor{headergray}
\multicolumn{11}{l}{\textbf{Target Domain: IUGC 24 Dataset \cite{bai2024intrapartum}}} \\
\midrule[1.5pt]
DSTCT~\cite{linguraru_intrapartum_2024} & \cellcolor{secondgray}$4.99_{\pm 4.70}$ & $7.33_{\pm 3.25}$ & $4.40_{\pm 3.04}$ & $8.52_{\pm 4.43}$ & $0.93_{\pm 0.03}$ & $0.89_{\pm 0.05}$ & $0.93_{\pm 0.03}$ & $28.57_{\pm 15.51}$ & $15.57_{\pm 27.21}$ & $27.35_{\pm 15.68}$ \\
IGT-Net~\cite{CHEN2026128998} & $6.81_{\pm 6.33}$ & $8.28_{\pm 3.42}$ & $5.04_{\pm 4.42}$ & $9.63_{\pm 4.57}$ & $0.91_{\pm 0.03}$ & $0.88_{\pm 0.06}$ & $0.92_{\pm 0.04}$ & $33.84_{\pm 16.27}$ & $19.67_{\pm 33.97}$ & $32.35_{\pm 16.71}$ \\
UpNet~\cite{BAI2025103353} & $31.63_{\pm 25.84}$ & $31.48_{\pm 6.41}$ & $28.66_{\pm 16.76}$ & $35.24_{\pm 8.22}$ & $0.69_{\pm 0.08}$ & $0.23_{\pm 0.19}$ & $0.72_{\pm 0.09}$ & $130.25_{\pm 29.39}$ & $66.98_{\pm 23.57}$ & $133.26_{\pm 31.81}$ \\
TransU-Net~\cite{BAI2025103353} & $16.18_{\pm 15.95}$ & $21.45_{\pm 8.24}$ & $13.89_{\pm 6.70}$ & $28.06_{\pm 10.82}$ & $0.75_{\pm 0.09}$ & $0.56_{\pm 0.19}$ & $0.76_{\pm 0.10}$ & $82.50_{\pm 31.71}$ & $42.87_{\pm 35.27}$ & $97.69_{\pm 30.29}$ \\
AoP-SAM~\cite{ZHOU2025125699} & $5.13_{\pm 4.68}$ & \cellcolor{secondgray}$3.26_{\pm 1.60}$ & \cellcolor{secondgray}$1.82_{\pm 0.98}$ & \cellcolor{secondgray}$3.80_{\pm 2.13}$ & \cellcolor{secondgray}$0.93_{\pm 0.03}$ & \cellcolor{secondgray}$0.90_{\pm 0.05}$ & \cellcolor{secondgray}$0.94_{\pm 0.03}$ & \cellcolor{bestblue}$12.74_{\pm 7.16}$ & \cellcolor{secondgray}$5.97_{\pm 3.37}$ & \cellcolor{bestblue}$12.23_{\pm 7.35}$ \\
USFM~\cite{JIAO2024103202} & $10.84_{\pm 9.91}$ & $5.21_{\pm 2.31}$ & $4.29_{\pm 3.86}$ & $5.82_{\pm 3.19}$ & $0.89_{\pm 0.05}$ & $0.79_{\pm 0.11}$ & $0.90_{\pm 0.05}$ & $19.26_{\pm 9.31}$ & $17.74_{\pm 27.83}$ & $17.26_{\pm 8.85}$ \\
\textbf{R2AoP (Ours)} & \cellcolor{bestblue}$\textbf{4.22}_{\pm 3.40}$ & \cellcolor{bestblue}$\textbf{2.32}_{\pm 1.11}$ & \cellcolor{bestblue}$\textbf{1.27}_{\pm 0.66}$ & \cellcolor{bestblue}$\textbf{2.70}_{\pm 1.47}$ & \cellcolor{bestblue}$\textbf{0.95}_{\pm 0.02}$ & \cellcolor{bestblue}$\textbf{0.93}_{\pm 0.03}$ & \cellcolor{bestblue}$\textbf{0.96}_{\pm 0.02}$ & \cellcolor{secondgray}$\textbf{12.81}_{\pm 9.16}$ & \cellcolor{bestblue}$\textbf{4.93}_{\pm 3.09}$ & \cellcolor{secondgray}$\textbf{12.95}_{\pm 10.04}$ \\
$\Delta$ Gain & \textcolor[HTML]{199935}{$-0.77$} & \textcolor[HTML]{199935}{$-0.94$} & \textcolor[HTML]{199935}{$-0.55$} & \textcolor[HTML]{199935}{$-1.10$} & \textcolor[HTML]{199935}{$+0.02$} & \textcolor[HTML]{199935}{$+0.03$} & \textcolor[HTML]{199935}{$+0.02$} & \textcolor[HTML]{199935}{$+0.07$} & \textcolor[HTML]{199935}{$-1.04$} & \textcolor[HTML]{199935}{$+0.72$} \\
\bottomrule[1.5pt]
\end{tabular}%
}
\end{table}

\begin{figure}[t]
    \centering
    \includegraphics[width=1.0\linewidth]{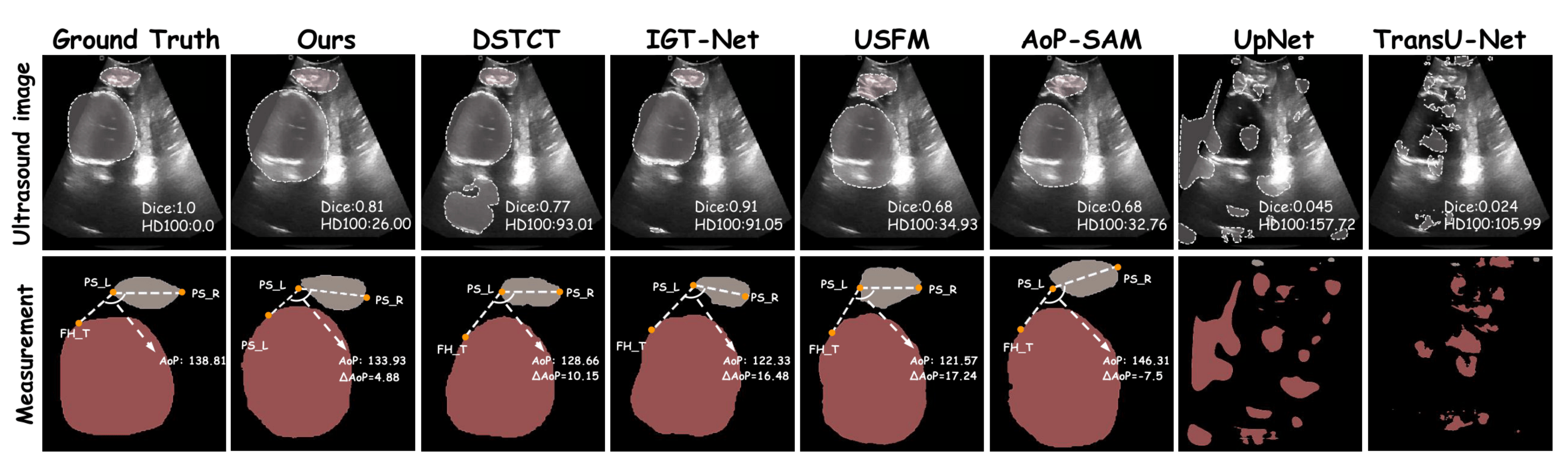}
    \caption{Qualitative comparison of PS/FH segmentation and AoP estimation across methods. The top row shows segmentation results, and the bottom row shows AoP geometry visualization including the PS axis and FH tangent.}
    \label{fig:contrast}
\end{figure}

\subsection{Experiment Results}
\textbf{Comparative Experiments.} As shown in Table~\ref{tab:comparison} and Fig.~\ref{fig:contrast}, the proposed method achieves the best performance on the JNU-IFM test set under challenging cross-domain acquisition variability. The AoP error is reduced to 6.21, while the overall ASD and HD100 reach 3.12 and 13.14, respectively, both lower than those of competing methods, indicating more accurate localization of critical structural boundaries. For overlap accuracy, the overall Dice score reaches 0.9232, maintaining stable performance across both PS and FH structures.

On the IUGC 2024 test set, the proposed method further demonstrates strong cross-domain generalization and achieves the best results across major evaluation metrics. The overall Dice score increases to 0.9525, ASD decreases to 2.32, and the AoP error is reduced to 4.22, substantially outperforming other methods. Overall, the results confirm robust generalization and clinical applicability.

\begin{table}[t]
\centering
\caption{Ablation study on JNU-IFM 23 and IUGC 24 datasets. The best and second-best results are highlighted in \colorbox{bestblue}{blue} and \colorbox{secondgray}{gray}, respectively. $\Delta$ Gain represents the performance range (best mean minus worst mean) across all ablation configurations, quantifying the contribution margin of each component.}
\label{tab:ablation}
\resizebox{\textwidth}{!}{%
\begin{tabular}{l!{\vrule width 1.5pt}c!{\vrule width 1.0pt}ccc!{\vrule width 1.0pt}ccc!{\vrule width 1.0pt}ccc}
\toprule[1.5pt]
\multicolumn{1}{c!{\vrule width 1.5pt}}{\multirow{2}{*}{\textbf{Method}}} & 
\multicolumn{1}{c!{\vrule width 1.0pt}}{\multirow{2}{*}{\textbf{AoP$\pm${std}}~$\downarrow$}} & 
\multicolumn{3}{c!{\vrule width 1.0pt}}{\cellcolor[HTML]{F0F0FF}\textbf{ASD$\pm${std}~(mm)}~$\downarrow$} & 
\multicolumn{3}{c!{\vrule width 1.0pt}}{\cellcolor[HTML]{E5F6FD}\textbf{Dice$\pm${std}}~$\uparrow$} & 
\multicolumn{3}{c}{\cellcolor[HTML]{F7F0F8}\textbf{HD100$\pm${std}~(mm)}~$\downarrow$} \\
\cmidrule[1.0pt]{3-5} \cmidrule[1.0pt]{6-8} \cmidrule[1.0pt]{9-11}
 &  & 
\textbf{PSFH} & \textbf{PS} & \textbf{FH} & 
\textbf{PSFH} & \textbf{PS} & \textbf{FH} & 
\textbf{PSFH} & \textbf{PS} & \textbf{FH} \\
\midrule[1.5pt]
\rowcolor{headergray}
\multicolumn{11}{l}{\textbf{Target Domain: JNU-IFM 23 Dataset \cite{LU2022107904}}} \\
\midrule[1.5pt]
w/o $\mathcal{L}_{\mathrm{aop}}$ & $6.48_{\pm 5.62}$ & \cellcolor{secondgray}$3.16_{\pm 2.44}$ & $2.27_{\pm 1.40}$ & $3.46_{\pm 3.17}$ & \cellcolor{secondgray}$0.92_{\pm 0.04}$ & $0.85_{\pm 0.08}$ & \cellcolor{secondgray}$0.93_{\pm 0.04}$ & $13.44_{\pm 14.55}$ & $7.37_{\pm 4.48}$ & $12.42_{\pm 14.78}$ \\
w/o $\mathcal{L}_{\mathrm{ent}}$ & $6.53_{\pm 5.57}$ & $3.27_{\pm 2.28}$ & $2.32_{\pm 1.36}$ & $3.60_{\pm 2.99}$ & $0.92_{\pm 0.04}$ & $0.84_{\pm 0.08}$ & $0.92_{\pm 0.05}$ & $13.59_{\pm 12.08}$ & $7.73_{\pm 4.03}$ & $12.51_{\pm 12.47}$ \\
w/o $\mathcal{L}_{\mathrm{tv}}$ & $6.59_{\pm 6.11}$ & $3.17_{\pm 2.11}$ & $2.43_{\pm 1.73}$ & \cellcolor{secondgray}$3.45_{\pm 2.74}$ & $0.92_{\pm 0.04}$ & $0.84_{\pm 0.11}$ & $0.93_{\pm 0.04}$ & $13.37_{\pm 11.42}$ & $7.74_{\pm 4.56}$ & \cellcolor{secondgray}$12.02_{\pm 11.47}$ \\
w/o Triple Branches & $6.61_{\pm 5.90}$ & $3.25_{\pm 2.15}$ & \cellcolor{secondgray}$2.25_{\pm 1.32}$ & $3.61_{\pm 2.82}$ & $0.92_{\pm 0.04}$ & \cellcolor{secondgray}$0.85_{\pm 0.08}$ & $0.93_{\pm 0.04}$ & $14.19_{\pm 14.98}$ & \cellcolor{secondgray}$7.17_{\pm 3.84}$ & $13.49_{\pm 15.09}$ \\
w/o Parameter Freezing & \cellcolor{secondgray}$6.45_{\pm 6.06}$ & $3.16_{\pm 2.46}$ & $2.34_{\pm 1.41}$ & \cellcolor{bestblue}$3.43_{\pm 3.15}$ & $0.92_{\pm 0.04}$ & $0.84_{\pm 0.09}$ & $0.93_{\pm 0.04}$ & \cellcolor{secondgray}$13.17_{\pm 11.48}$ & $7.74_{\pm 4.13}$ & \cellcolor{bestblue}$11.78_{\pm 11.69}$ \\
\textbf{Ours} & \cellcolor{bestblue}$\textbf{6.21}_{\pm 5.18}$ & \cellcolor{bestblue}$\textbf{3.12}_{\pm 2.88}$ & \cellcolor{bestblue}$\textbf{2.00}_{\pm 1.20}$ & $\textbf{3.49}_{\pm 3.62}$ & \cellcolor{bestblue}$\textbf{0.92}_{\pm 0.04}$ & \cellcolor{bestblue}$\textbf{0.86}_{\pm 0.08}$ & \cellcolor{bestblue}$\textbf{0.93}_{\pm 0.04}$ & \cellcolor{bestblue}$\textbf{13.14}_{\pm 13.06}$ & \cellcolor{bestblue}$\textbf{6.76}_{\pm 4.04}$ & $\textbf{12.28}_{\pm 13.18}$ \\
\rowcolor{white}
$\Delta$ Gain & \textcolor[HTML]{199935}{$-0.40$} & \textcolor[HTML]{199935}{$-0.15$} & \textcolor[HTML]{199935}{$-0.43$} & \textcolor[HTML]{199935}{$-0.18$} & \textcolor[HTML]{199935}{$0.00$} & \textcolor[HTML]{199935}{$+0.02$} & \textcolor[HTML]{199935}{$+0.01$} & \textcolor[HTML]{199935}{$-1.05$} & \textcolor[HTML]{199935}{$-0.98$} & \textcolor[HTML]{199935}{$-1.51$} \\
\midrule[1.5pt]
\rowcolor{headergray}
\multicolumn{11}{l}{\textbf{Target Domain: IUGC 24 Dataset \cite{bai2024intrapartum}}} \\
\midrule[1.5pt]
w/o $\mathcal{L}_{\mathrm{aop}}$ & \cellcolor{secondgray}$5.16_{\pm 5.36}$ & $3.83_{\pm 1.74}$ & $2.34_{\pm 1.21}$ & $4.38_{\pm 2.26}$ & $0.92_{\pm 0.03}$ & $0.87_{\pm 0.06}$ & $0.93_{\pm 0.03}$ & $15.32_{\pm 8.59}$ & $7.22_{\pm 3.99}$ & $14.77_{\pm 8.88}$ \\
w/o $\mathcal{L}_{\mathrm{ent}}$ & $5.78_{\pm 5.67}$ & $3.58_{\pm 1.83}$ & $1.97_{\pm 1.03}$ & $4.17_{\pm 2.43}$ & $0.93_{\pm 0.03}$ & $0.89_{\pm 0.05}$ & $0.93_{\pm 0.04}$ & $15.59_{\pm 9.46}$ & $6.48_{\pm 3.56}$ & $15.14_{\pm 9.68}$ \\
w/o $\mathcal{L}_{\mathrm{tv}}$ & $5.57_{\pm 5.70}$ & $3.59_{\pm 1.86}$ & $1.91_{\pm 0.97}$ & $4.20_{\pm 2.46}$ & \cellcolor{secondgray}$0.93_{\pm 0.03}$ & $0.90_{\pm 0.05}$ & \cellcolor{secondgray}$0.93_{\pm 0.03}$ & $15.44_{\pm 9.28}$ & \cellcolor{secondgray}$6.38_{\pm 3.51}$ & $15.01_{\pm 9.58}$ \\
w/o Triple Branches & $5.20_{\pm 5.27}$ & $3.57_{\pm 1.77}$ & \cellcolor{secondgray}$1.89_{\pm 1.00}$ & $4.19_{\pm 2.33}$ & $0.93_{\pm 0.03}$ & \cellcolor{secondgray}$0.90_{\pm 0.05}$ & $0.93_{\pm 0.03}$ & \cellcolor{secondgray}$14.35_{\pm 8.66}$ & $6.46_{\pm 3.31}$ & \cellcolor{secondgray}$13.83_{\pm 8.89}$ \\
w/o Parameter Freezing & $5.55_{\pm 5.48}$ & \cellcolor{secondgray}$3.56_{\pm 1.73}$ & $1.94_{\pm 1.05}$ & \cellcolor{secondgray}$4.16_{\pm 2.30}$ & $0.93_{\pm 0.03}$ & $0.89_{\pm 0.05}$ & $0.93_{\pm 0.03}$ & $15.31_{\pm 8.89}$ & $6.45_{\pm 3.52}$ & $14.88_{\pm 9.10}$ \\
\textbf{Ours} & \cellcolor{bestblue}$\textbf{4.22}_{\pm 3.40}$ & \cellcolor{bestblue}$\textbf{2.32}_{\pm 1.11}$ & \cellcolor{bestblue}$\textbf{1.27}_{\pm 0.66}$ & \cellcolor{bestblue}$\textbf{2.70}_{\pm 1.47}$ & \cellcolor{bestblue}$\textbf{0.95}_{\pm 0.02}$ & \cellcolor{bestblue}$\textbf{0.93}_{\pm 0.03}$ & \cellcolor{bestblue}$\textbf{0.96}_{\pm 0.02}$ & \cellcolor{bestblue}$\textbf{12.81}_{\pm 9.16}$ & \cellcolor{bestblue}$\textbf{4.93}_{\pm 3.09}$ & \cellcolor{bestblue}$\textbf{12.95}_{\pm 10.04}$ \\
\rowcolor{white}
$\Delta$ Gain & \textcolor[HTML]{199935}{$-1.56$} & \textcolor[HTML]{199935}{$-1.51$} & \textcolor[HTML]{199935}{$-1.07$} & \textcolor[HTML]{199935}{$-1.68$} & \textcolor[HTML]{199935}{$+0.03$} & \textcolor[HTML]{199935}{$+0.06$} & \textcolor[HTML]{199935}{$+0.03$} & \textcolor[HTML]{199935}{$-2.78$} & \textcolor[HTML]{199935}{$-2.29$} & \textcolor[HTML]{199935}{$-2.19$} \\
\bottomrule[1.5pt]
\end{tabular}%
}
\end{table}

\begin{figure}[!t]
    \centering
    \includegraphics[width=1.0\linewidth]{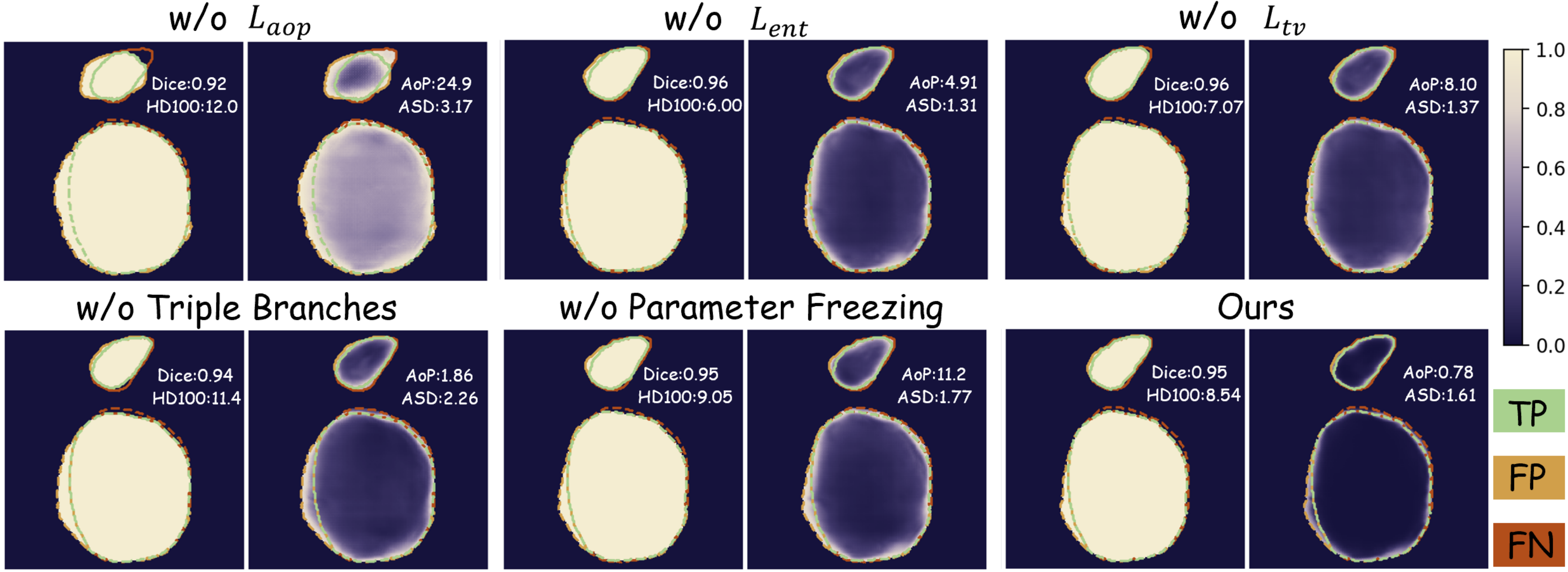}
    \caption{Qualitative ablation results of different components in the proposed method. The left panels show segmentation error maps with TP, FP, and FN, and the right panels show confidence maps with AoP error and boundary metrics.}
    \label{fig:ablation}
\end{figure}

\textbf{Ablation Study.} As shown in Table~\ref{tab:ablation} and Fig.~\ref{fig:ablation}, removing the three-branch gating mechanism causes consistent degradation across both target domains, with increased AoP error, HD100, and ASD, indicating its role in key boundary discrimination. Eliminating the geometry reliability constraint has the largest impact on the IUGC 2024 dataset, where PS boundary accuracy and AoP stability decrease. Removing entropy minimization and TV smoothing increases AoP error and HD100, while removing the lightweight update strategy further amplifies cross-domain degradation. Overall, these components complement each other in structural discrimination, boundary reliability, and cross-domain robustness, jointly improving AoP estimation.

\section{Conclusion}
This study establishes a geometry-aware paradigm for reliable AoP estimation, explicitly addressing the long-standing instability caused by error amplification in cascaded segmentation–measurement pipelines. By coupling boundary reliability with geometric computation, the proposed framework enables stable and interpretable AoP quantification under challenging ultrasound conditions. Beyond performance gains on multi-center benchmarks, our findings suggest that enforcing geometric reliability is critical for making automated obstetric measurements clinically usable. The proposed formulation is general and can be extended to other geometry-dependent measurements in ultrasound and beyond, providing a principled direction for robust clinical deployment.

\begin{credits}
\subsubsection{Acknowledgements} This work was supported by the National Natural Science Foundation of China (No. 82302166), Tsinghua University Startup Fund, Fundamental Research Funds for the Provincial Universities of Zhejiang (No. GK259909299001-006). 

\subsubsection{Disclosure of Interests} The authors declare no competing interests.
\end{credits}

%
%
%
%

\bibliographystyle{splncs04}
\bibliography{ref}

\begin{thebibliography}{10}
\providecommand{\url}[1]{\texttt{#1}}
\providecommand{\urlprefix}{URL }
\providecommand{\doi}[1]{https://doi.org/#1}

\bibitem{BAI2026103960}
Iugc: A benchmark of landmark detection in end-to-end intrapartum ultrasound biometry. Medical Image Analysis  \textbf{110},  103960 (2026)

\bibitem{bai2024intrapartum}
Bai, J., Lekadir, K., Ni, D., Slimani, S., Campello, V.M., Ohene-Botwe, B., Lu, Y., Chen, G., Hou, H., Qiu, D., Zhou, Z.: Intrapartum ultrasound grand challenge 2024 (2024)

\bibitem{BAI2025103353}
Bai, J., Zhou, Z., Ou, Z., Koehler, G., Stock, R., Maier-Hein, K., Elbatel, M., Mart{\'i}, R., Li, X., Qiu, Y., Gou, P., Chen, G., Zhao, L., Zhang, J., Dai, Y., Wang, F., Silvestre, G., Curran, K., Sun, H., Xu, J., Cai, P., Jiang, L., Lan, L., Ni, D., Zhong, M., Chen, G., Campello, V.M., Lu, Y., Lekadir, K.: {PSFHS} challenge report: Pubic symphysis and fetal head segmentation from intrapartum ultrasound images. Medical Image Analysis  \textbf{99},  103353 (2025)

\bibitem{chen2024psfhs}
Chen, G., Bai, J., Ou, Z., Lu, Y., Wang, H.: {PSFHS}: intrapartum ultrasound image dataset for {AI}-based segmentation of pubic symphysis and fetal head. Scientific Data  \textbf{11}(1), ~436 (2024)

\bibitem{10.1145/3696409.3700214}
Chen, S., Wang, H., Long, S., Bai, J., Jiang, J.: Ultrasound video segmentation of pubic symphysis and fetal head for {{AoP}} measurement. In: Proceedings of the 6th ACM International Conference on Multimedia in Asia. {{MMAsia}} '24, New York, NY, USA (2024)

\bibitem{CHEN2024107917}
Chen, Y., Zhang, C., Chen, B., Huang, Y., Sun, Y., Wang, C., Fu, X., Dai, Y., Qin, F., Peng, Y., Gao, Y.: Accurate leukocyte detection based on deformable-{{DETR}} and multi-level feature fusion for aiding diagnosis of blood diseases. Computers in Biology and Medicine  \textbf{170},  107917 (2024)

\bibitem{chen2024dualpath}
Chen, Z., Lu, Y., Long, S., Campello, V.M., Bai, J., Lekadir, K.: Fetal head and pubic symphysis segmentation in intrapartum ultrasound image using a dual-path boundary-guided residual network. IEEE Journal of Biomedical and Health Informatics  \textbf{28}(8),  4648--4659 (2024)

\bibitem{CHEN2026128998}
Chen, Z., Ou, Z., Lu, Y., Campello, V.M., Bai, J., Lekadir, K.: Uncertainty-fetal head and pubic symphysis segmentation with enhanced multi-scale features and sparse visual graph attention. Expert Systems with Applications  \textbf{296},  128998 (2026)

\bibitem{10.1007/978-3-032-11616-1_8}
Deng, B., Chen, Y., Peng, Z.: A two-stage semi-supervised ensemble framework for automated angle of progression measurement in intrapartum ultrasound. In: Bai, J., Huang, Y., Khobo, I., Yaqub, M., Lekadir, K., Ni, D., Li, S. (eds.) Intrapartum Ultrasound. pp. 88--99. Cham (2026)

\bibitem{https://doi.org/10.1002/uog.7521}
Dückelmann, A.M., Bamberg, C., Michaelis, S.A.M., Lange, J., Nonnenmacher, A., Dudenhausen, J.W., Kalache, K.D.: Measurement of fetal head descent using the ‘angle of progression’ on transperineal ultrasound imaging is reliable regardless of fetal head station or ultrasound expertise. Ultrasound in Obstetrics \& Gynecology  \textbf{35}(2),  216--222 (2010)

\bibitem{10.1007/978-3-031-96318-6_3}
Gan, J., Liang, Z., Fan, J., Mcguire, L., Clarke, J., Cai, W.: Accurate fetal head descent assessment during labor using video {{Swin Transformer}} and wavelet-based multitask learning for 2024 {{MICCAI}} challenge {{IUGC}}. In: Bai, J., Lu, Y. (eds.) Intrapartum Ultrasound. pp. 21--31. Cham (2026)

\bibitem{linguraru_intrapartum_2024}
Jiang, J., Wang, H., Bai, J., Long, S., Chen, S., Campello, V.M., Lekadir, K.: Intrapartum {Ultrasound} {Image} {Segmentation} of {Pubic} {Symphysis} and {Fetal} {Head} {Using} {Dual} {Student}-{Teacher} {Framework} with {{CNN}}-{{ViT}} {Collaborative} {Learning}. In: Linguraru, M.G., Dou, Q., Feragen, A., Giannarou, S., Glocker, B., Lekadir, K., Schnabel, J.A. (eds.) Medical {Image} {Computing} and {Computer} {Assisted} {Intervention} – {{MICCAI}} 2024, vol. 15001, pp. 448--458. Cham (2024)

\bibitem{JIAO2024103202}
Jiao, J., Zhou, J., Li, X., Xia, M., Huang, Y., Huang, L., Wang, N., Zhang, X., Zhou, S., Wang, Y., Guo, Y.: {USFM}: A universal ultrasound foundation model generalized to tasks and organs towards label efficient image analysis. Medical Image Analysis  \textbf{96},  103202 (2024)

\bibitem{10.1007/978-3-032-11616-1_1}
Liu, X., Hu, J., Li, Y., Chen, X., Wang, Y.: Noisy student-based self-training enhances landmark detection in intrapartum ultrasound. In: Bai, J., Huang, Y., Khobo, I., Yaqub, M., Lekadir, K., Ni, D., Li, S. (eds.) Intrapartum Ultrasound. pp. 1--13. Cham (2026)

\bibitem{LU2022107904}
Lu, Y., Zhou, M., Zhi, D., Zhou, M., Jiang, X., Qiu, R., Ou, Z., Wang, H., Qiu, D., Zhong, M., Lu, X., Chen, G., Bai, J.: The {{JNU-IFM}} dataset for segmenting pubic symphysis-fetal head. Data in Brief  \textbf{41},  107904 (2022)

\bibitem{Ma_TTGA_2026}
Ma, X., Tao, Y., et~al.: Test-time generative augmentation for medical image segmentation. Medical Image Analysis  \textbf{109},  103902 (2026)

\bibitem{OU2024108501}
Ou, Z., Bai, J., Chen, Z., Lu, Y., Wang, H., Long, S., Chen, G.: {RTSeg-net}: A lightweight network for real-time segmentation of fetal head and pubic symphysis from intrapartum ultrasound images. Computers in Biology and Medicine  \textbf{175},  108501 (2024)

\bibitem{Ravishankar_IEEETMI_2025}
Ravishankar, H., Paluru, N., Sudhakar, P., Yalavarthy, P.K.: {Information Geometric} approaches for patient-specific test-time adaptation of deep learning models for semantic segmentation. IEEE Transactions on Medical Imaging  \textbf{44}(6),  2553--2567 (2025)

\bibitem{10.1007/978-3-031-96318-6_2}
Tan, Y., Wang, S., Shen, W.D., Yu, W.W., Li, Y., Zhang, P., Jiang, W., Li, Y.J.: Classification and segmentation of intrapartum ultrasound images with deep learning models. In: Bai, J., Lu, Y. (eds.) Intrapartum Ultrasound. pp. 11--20. Cham (2026)

\bibitem{10.1007/978-3-032-11616-1_10}
Tang, Y., Zhou, Z., Lu, Y., Bai, J., Long, S., Huang, Y., Khobo, I., Zhang, S., Zhou, Z., Guo, L.: Heatmap regression for automated angle of progression measurement: The baseline method for the {{IUGC2025}}. In: Intrapartum Ultrasound: {{MICCAI}} 2025 Grand Challenge, {{IUGC}} 2025, Held in Conjunction with {{MICCAI}} 2025, Daejeon, South Korea, September 23, 2025, Proceedings. pp. 105--117. Berlin, Heidelberg (2026)

\bibitem{wang2025smart}
Wang, Y., Chen, Y., Jiang, S., Yu, W., Liu, M., Wu, B., Zong, J., Qin, F., Wang, C., Tian, Q.: {SmaRT}: Style-modulated robust test-time adaptation for cross-domain brain tumor segmentation in {{MRI}}. arXiv preprint arXiv:2509.17925  (2025)

\bibitem{10.1007/978-3-031-96318-6_4}
Xia, Z., Li, H., Lan, L.: {{DSSAU-Net}}: U-shaped hybrid network for pubic symphysis and fetal head segmentation. In: Bai, J., Lu, Y. (eds.) Intrapartum Ultrasound. pp. 32--45. Cham (2026)

\bibitem{Benchmark2025}
Yu, W., Jiang, S., Chen, Y., Chang, S., Wang, Y., Wu, B., Dong, J., Liu, M., Zhu, S., Qin, F., Wang, C., Tian, Q.: A large scale benchmark for test time adaptation methods in medical image segmentation. arXiv preprint arXiv:2512.02497  (2025)

\bibitem{ZHOU2025125699}
Zhou, Z., Lu, Y., Bai, J., Campello, V.M., Feng, F., Lekadir, K.: {Segment Anything Model} for fetal head-pubic symphysis segmentation in intrapartum ultrasound image analysis. Expert Systems with Applications  \textbf{263},  125699 (2025)

\bibitem{ZHU2026104108}
Zhu, S., Chen, Y., Chen, W., Jiang, S., Zhou, G., Wang, Y., Qin, F., Wang, C., Tian, Q.: No modality left behind: Adapting to missing modalities via knowledge distillation for brain tumor segmentation. Medical Image Analysis  \textbf{112},  104108 (2026)

\bibitem{ZhuShe_Bridging_MICCAI2025}
Zhu, S., Chen, Y., Chen, W., Wang, Y., Liu, C., Jiang, S., Qin, F., Wang, C.: { Bridging the Gap in Missing Modalities: Leveraging Knowledge Distillation and Style Matching for Brain Tumor Segmentation }. In: proceedings of Medical Image Computing and Computer Assisted Intervention -- MICCAI 2025. vol. LNCS 15967, pp. 95 -- 106 (2025)

\end{thebibliography}
\end{document}